\documentclass[12pt,journal,compsoc,]{IEEEtran}

\makeatletter
\def\thickhline{%
	\noalign{\ifnum0=`}\fi\hrule \@height \thickarrayrulewidth \futurelet
	\reserved@a\@xthickhline}
\def\@xthickhline{\ifx\reserved@a\thickhline
	\vskip\doublerulesep
	\vskip-\thickarrayrulewidth
	\fi
	\ifnum0=`{\fi}}
\makeatother

\newlength{\thickarrayrulewidth}
\usepackage{epigraph}
\usepackage{longtable}

\usepackage{textcomp}

\usepackage{amssymb}

\usepackage[hidelinks]{hyperref}
\hypersetup{
	colorlinks=true,
	linkcolor=black,
	filecolor=black,
	citecolor=black,
	urlcolor=blue,
}

\usepackage{multirow}

\usepackage{color}
\usepackage[table, dvipsnames]{xcolor}
\definecolor{PyOrange}{RGB}{255, 201, 14}
\definecolor{PyBlue}{RGB}{112, 146, 190}

\definecolor{WordGreen}{RGB}{100, 136, 40}
\definecolor{WordDarkGrey}{RGB}{82, 82, 82}
\definecolor{WordRed}{RGB}{192, 80, 77}
\definecolor{WordBlue}{RGB}{0, 122, 192}

\definecolor{WordLightBlue}{RGB}{218, 238, 243}
\definecolor{WordLightGreen}{RGB}{234, 241, 221}

\definecolor{WordFillGreen}{RGB}{194, 214, 155}
\definecolor{WordFillRed}{RGB}{252, 214, 182}
\definecolor{WordFillGray}{RGB}{217, 217, 217}

\UseRawInputEncoding
\usepackage{times}
\usepackage{epsfig}
\usepackage{amssymb}
\usepackage[ruled,vlined,linesnumbered]{algorithm2e}
\usepackage{comment}
\usepackage{array, makecell} 
\usepackage{lettrine}
\usepackage{url} 
\usepackage{lscape}
\usepackage{tabularx}
\usepackage{colortbl}
\usepackage{hhline}
\usepackage{vcell}
\usepackage{longtable}
\usepackage{vcell}
\usepackage{rotating}
\usepackage{pdflscape}
\usepackage{sidecap}
\usepackage{neuralnetwork}
\usepackage[export]{adjustbox} % https://tex.stackexchange.com/questions/91566/syntax-similar-to-centering-for-right-and-left
\usepackage{acronym}
\acrodef{FCN}[FCN]{Fully Convolutional Network}
\acrodef{GAME}[GAME]{Grid Average Mean Absolute Error}
\acrodef{DL}[DL]{Deep Learning}
\acrodef{DNN}[DNN]{Deep Neural Network}
\acrodef{ML}[ML]{Machine Learning}
\acrodef{CV}[CV]{Computer Vision}
\acrodef{AI}[AI]{Artificial Intelligence}
\acrodef{CNN}[CNN]{Convolutional Neural Network}
\acrodef{RNN}[RNN]{Recurrent Neural Network}
\acrodef{GAN}[GAN]{Generative Adversarial Network}
\acrodef{JCU}[JCU]{James Cook University}
\acrodef{MAE}[MAE]{Mean Average Error}
\acrodef{MAP}[mAP]{Mean Average Precision}
\acrodef{CA}[CA]{Classification Accuracy}
\acrodef{LCFCN}[LCFCN]{Localization-based Counting loss Fully Convolutional Network}
\acrodef{IoT}[IoT]{Internet of Things}
\acrodef{MLP}[MLP]{Multi-Layer Perceptrons}
\acrodef{RAFT}[RAFT]{Recurrent All-Pairs Field Transforms }
\usepackage{xspace}

\usepackage[capitalize]{cleveref}
\crefname{section}{Sec.}{Section}
\usepackage{booktabs}
\usepackage{arydshln}
\usepackage{listings}
\definecolor{codegreen}{rgb}{0,0.6,0}
\definecolor{codegray}{rgb}{0.5,0.5,0.5}
\definecolor{codepurple}{rgb}{0.58,0,0.82}
\definecolor{backcolour}{rgb}{0.95,0.95,0.92}
\definecolor{cleacolorr}{rgb}{1,1,1}

\lstdefinestyle{mystyle}{
    backgroundcolor=\color{cleacolorr},   
    commentstyle=\color{codegreen},
    keywordstyle=\color{magenta},
    numberstyle=\tiny\color{codegray},
    stringstyle=\color{codepurple},
    basicstyle=\ttfamily\footnotesize,
    breakatwhitespace=false,         
    breaklines=true,                 
    captionpos=b,                    
    keepspaces=true,                 
    numbers=left,                    
    numbersep=5pt,                  
    showspaces=false,                
    showstringspaces=false,
    showtabs=false,                  
    tabsize=2
}

\usepackage[most]{tcolorbox}
\newtcolorbox[auto counter]{pabox}[2][]{%
colback=blue!5!white,colframe=blue!75!black,fonttitle=\bfseries,
title=Box~\thetcbcounter: #2,#1}
\newcommand{\MyPaperTitle}{How to Track and Segment Fish without Human Annotations: A Self-Supervised Deep Learning Approach}

\usepackage{tikz}

\ifCLASSOPTIONcompsoc
  \usepackage[caption=false,font=normalsize,labelfont=sf,textfont=sf]{subfig}
\else
  \usepackage[caption=false,font=footnotesize]{subfig}
\fi

\usepackage[numbers,sort&compress]{natbib}
\usepackage{graphicx}
\graphicspath{{./Graphics/}}
\DeclareGraphicsExtensions{.pdf,.jpeg,.png,.jpg}

\usepackage{amsmath}

\usepackage{array}

\usepackage{url}
\usepackage[switch]{lineno}

\begin{document}
% \linenumbers
\title{\MyPaperTitle}

% \author{
% 	Alzayat Saleh\textsuperscript{\href{https://orcid.org/0000-0001-6973-019X}{\includegraphics[scale=0.035]{Logo_ORCID.png}}},~\IEEEmembership{Graduate Student Member,~IEEE,} 
% 	David B. Jones\textsuperscript{\href{https://orcid.org/0000-0001-6973-019X}{\includegraphics[scale=0.035]{Logo_ORCID.png}}},\\
% 	Dean R. Jerry\textsuperscript{\href{https://orcid.org/0000-0001-6973-019X}{\includegraphics[scale=0.035]{Logo_ORCID.png}}},
%     and Mostafa~Rahimi~Azghadi\textsuperscript{\href{https://orcid.org/0000-0001-7975-3985}{\includegraphics[scale=0.035]{Logo_ORCID.png}}},~\IEEEmembership{Senior Member,~IEEE}%
% }
% \author{
% 	Alzayat Saleh\textsuperscript{\href{https://orcid.org/0000-0001-6973-019X}{\includegraphics[scale=0.035]{Logo_ORCID.png}}}, 
% 	Marcus Sheaves\textsuperscript{\href{https://orcid.org/0000-0003-0662-3439}{\includegraphics[scale=0.035]{Logo_ORCID.png}}},
% 	Dean Jerry\textsuperscript{\href{https://orcid.org/0000-0003-3735-1798}{\includegraphics[scale=0.035]{Logo_ORCID.png}}}, 
%     and Mostafa~Rahimi~Azghadi\textsuperscript{\href{https://orcid.org/0000-0001-7975-3985}{\includegraphics[scale=0.035]{Logo_ORCID.png}}}
% }

%%%%%%% Blinded Manuscript %%%%%%%%%%%%%%%%%%%%%
\author{
    \IEEEauthorblockN{Alzayat Saleh\IEEEauthorrefmark{1}, Marcus Sheaves\IEEEauthorrefmark{1}, 
    Dean Jerry\IEEEauthorrefmark{1}\IEEEauthorrefmark{2}, 
    and Mostafa~Rahimi~Azghadi\IEEEauthorrefmark{1}\IEEEauthorrefmark{2}\IEEEauthorrefmark{3}}
    
    \IEEEauthorblockA{\IEEEauthorrefmark{1}College of Science and Engineering, James Cook University, Townsville, QLD, Australia}
    
    \IEEEauthorblockA{\IEEEauthorrefmark{2}ARC Research Hub for Supercharging Tropical Aquaculture through Genetic Solutions, James Cook University, Townsville, QLD, Australia}
    \IEEEauthorblockA{\IEEEauthorrefmark{3}Corresponding author: mostafa.rahimiazghadi@jcu.edu.au}
}
%%%%%%% Blinded Manuscript %%%%%%%%%%%%%%%%%%%%%

% \affil[1]{College of Science and Engineering, James Cook University, Townsville, QLD, Australia}
% \affil[2]{ARC Research Hub for Supercharging Tropical Aquaculture through Genetic Solutions, James Cook University, Townsville, QLD, Australia}

\maketitle

\begin{abstract} Tracking fish movements and sizes of fish is crucial to understanding their ecology and behaviour. Knowing where fish migrate, how they interact with their environment, and how their size affects their behaviour can help ecologists develop more effective conservation and management strategies to protect fish populations and their habitats. Deep learning is a promising tool to analyze fish ecology from underwater videos. However, training deep neural networks (DNNs) for fish tracking and segmentation requires high-quality labels, which are expensive to obtain. We propose an alternative unsupervised approach that relies on spatial and temporal variations in video data to generate noisy pseudo-ground-truth labels. We train a multitask DNN using these pseudo-labels. Our framework consists of three stages: (1) an optical flow model generates the pseudo labels using spatial and temporal consistency between frames, (2) a self-supervised model refines the pseudo-labels incrementally, and (3) a segmentation network uses the refined labels for training. Consequently, we perform extensive experiments to validate our method on three public underwater video datasets and demonstrate its effectiveness for video annotation and segmentation. We also evaluate its robustness to different imaging conditions and discuss its limitations. \end{abstract}

\ifCLASSOPTIONpeerreview
\else
\fi

%%%%%%%%%%%%%%%%%%%%%%%%%%%%%%%%%%%%%%%%%%%%%%%%%%%%%%%%%%%%%%%%
%%%%%%%%%%%%%%%%%%%%%%%%%%%%%%%%%%%%%%%%%%%%%%%%%%%%%%%%%%%%%%%%
\section{Introduction}\label{secintro}

The automatic tracking and segmentation of individual fish have emerged as pivotal tools in the field of ecological behavioural analysis, with a broad spectrum of applications. This is evidenced by numerous studies in the domain \cite{lopiz2021eco,zou2021fusion, gatti2021track, wageeh2021yolo, saleh2022a, zhang2023uvosam}.
The ability to understand and predict animal motion in their natural habitats could yield significant benefits across various research and industry domains \cite{guida2013eco, Zebrafish2019zebra, olsen2012cod, wang2012wild, dutta2023robust}. However, the inherent complexity of animal movement in the wild presents a great challenge. Factors contributing to this complexity include intermittent visibility of animals in videos and the presence of multiple animals within a single video frame, both of which complicate tracking and segmentation tasks. Addressing these challenges often necessitates the deployment of advanced computational methods.

A large number of studies have attempted to tackle these challenges \cite{javed2022visual, Saleh2020, Konovalov2019, Laradji2021, Konovalov2018, Konovalov2019a, JAHANBAKHT2023102303}. These studies predominantly rely on pixel-level annotations to train or enhance their Deep Neural Networks (DNNs). However, obtaining these annotations is both costly and time-consuming, particularly for fish segmentation in the wild.
Most current automated methods operate under the assumption that training data is typically paired with ground truth derived from videos containing a large number of fish \cite{Wang2017b, Villon2018a, Saleh2020, li2021tracking, cao2022track}. Despite the high cost associated with obtaining ground truth, it is necessary to acquire a substantial number of video sequences. This is due to the difficulty in achieving accurate results using only a limited number of sample videos.

\begin{figure}[!h!t]
\centering
\includegraphics[width=0.98\linewidth]{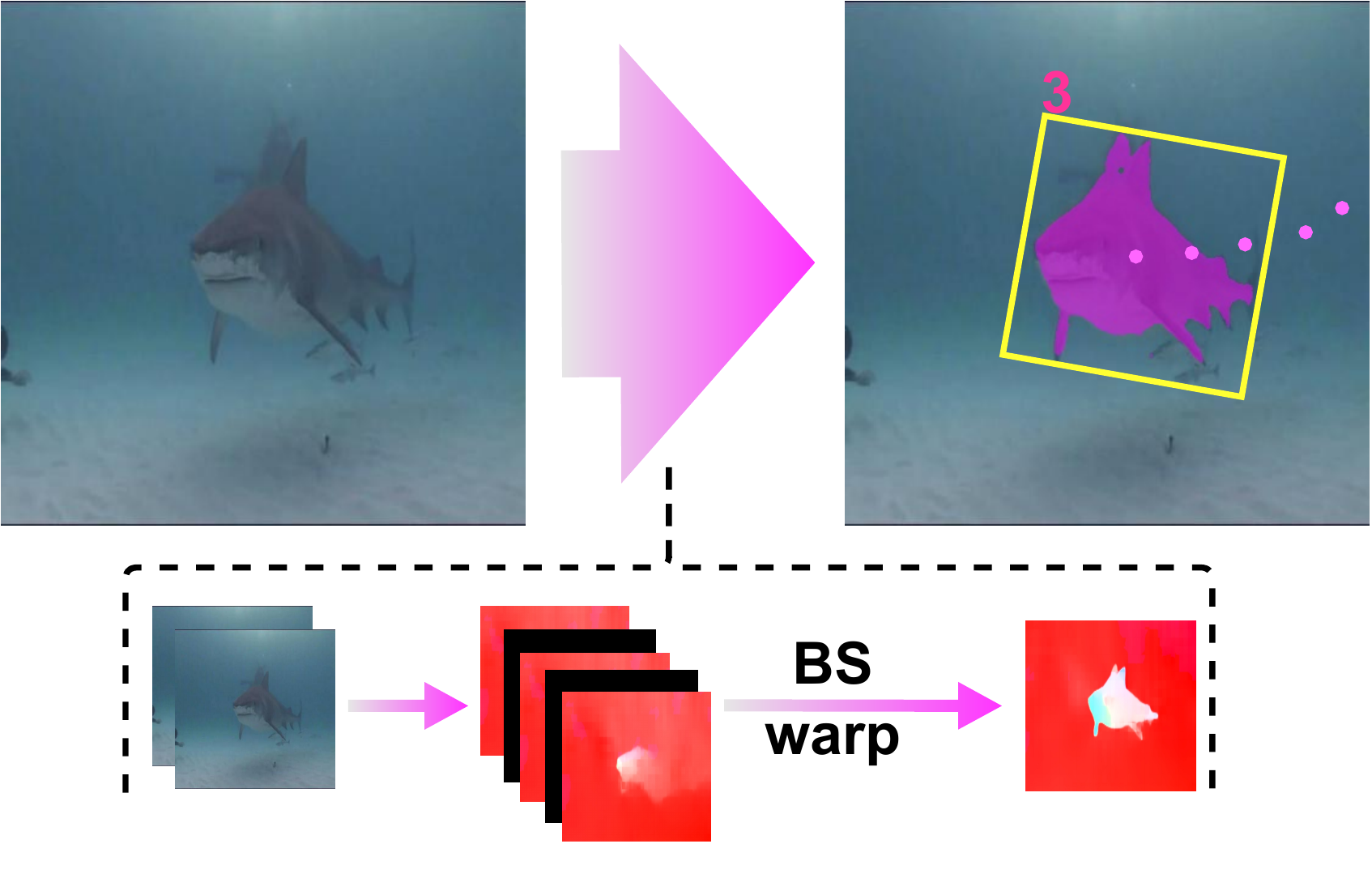}
\caption{Combining background subtraction and optical flow demonstrate how both levels work in concert to preserve object boundaries and temporal coherence throughout the video. 
% See \cref{fig:4} and comparison video  at\\
% \href{https://youtu.be/8LOKsVSiY9U}{https://youtu.be/8LOKsVSiY9U}. 
Please refer to \cref{secfram} for details.}
\label{fig:10}
\end{figure}
% %%%%%%%%%%%%%%%%%%%%%%%%%%%%%%%%%%%%%%%%%%%%%%%%%%%%%%%%%%%%%%%%

This study was motivated by the importance of the challenges faced when trying to annotate and segment animals in videos in the wild. Unlike in controlled conditions, where animals are easily distinguishable from the background, fish are difficult to distinguish in realistic videos \cite{Ditria2021a, Saleh2020a, proencca2023trade}, even with domain knowledge. This is due to the large variations in the appearance of the animals, lighting conditions, and background. 

Our approach aims to develop an unsupervised method for fish tracking and segmentation without the need for human annotations, by leveraging spatial and temporal variations in video data using known techniques of background subtraction and optical flow, as shown in \cref{fig:10}.
Specifically,  we propose to generate pseudo labels based on unlabelled video data. The use of pseudo labels can benefit various learning-based algorithms since it can significantly reduce the labelling cost. The key to the proposed method is to take advantage of the intrinsic temporal consistency between consecutive frames to improve the generated labels by refining them with a self-supervised model. We propose training a \ac{DNN} to segment individual fish based on the generated pseudo-labels. As long as the pseudo-labels are generated in a way that they have similar structure and appearance to real ones, the model can learn to understand the underlying structure from the pseudo-labels. 
In general, the more realistic the pseudo-labels, the better the segmentation accuracy.
We include a short video of our model prediction in \href{https://youtu.be/Z5G7YBoL3eM}{https://youtu.be/Z5G7YBoL3eM}
and \href{https://youtu.be/8LOKsVSiY9U}{https://youtu.be/8LOKsVSiY9U}.

The main contributions of this paper are listed as follows:
\begin{itemize}
    \item Propose to use pixel-level pseudo labels generated by an optical-flow model and background subtraction to learn the segmentation and tracking of individual fish automatically without manual interaction. 
    \item Demonstrate that using self-supervised refinement, we can further improve the accuracy of the pseudo labels for fish tracking and segmentation.
    \item Evaluate our method on three public datasets with different image quality. 
    \item Discuss the limitations of the current model and our future research directions.
\end{itemize}

\noindent  The rest of the paper is organized as follows. \cref{secrlt} covers related works and provides background information on the novel aspects of our work. Our model's framework is described in detail in  \cref{secfram}. \cref{secmethod}  presents our method for training and evaluating our model. The experimental setup and results are presented in  \cref{secresult}, while detailed discussions of our results are presented in  \cref{secdisc}. Finally, \cref{secconc} concludes our paper.

\section{Related Work} \label{secrlt}

The field of video object segmentation and animal tracking has witnessed substantial advancements in recent years. A noteworthy contribution is the unsupervised video object segmentation model, UVOSAM, proposed by Zhang et al. \cite{zhang2023uvosam}. This model, which operates without the need for masks, has introduced new possibilities in the field.
In the realm of animal tracking, Dutta et al. \cite{dutta2023robust} have developed a deep learning workflow that holds particular relevance for ecological studies. Complementing this, Javed et al. \cite{javed2022visual}   have provided a comprehensive survey of visual object tracking techniques, significantly enhancing our understanding of the current landscape.

Further enriching the field, Cao et al. \cite{cao2022track}  proposed a method of dense spatio-temporal position encoding to improve tracking accuracy. Proencca et al. \cite{proencca2023trade} introduced TRADE, a method that utilises 3D trajectory and ground depth estimates for UAVs.
In terms of segmentation, Jahanbakht et al. \cite{jahanbakht2022distributed} explored distributed deep learning and energy-efficient image processing for fish segmentation. Zhang et al. \cite{zhang10msgnet}  developed MSGNet, which uses multiple sources of information to improve the precision of fish segmentation. This approach has been influential in shaping our own methodology.
In the following subsection, we will provide a brief review of the research domains that are most relevant to our work.

% Recent years have seen significant progress in the field of video object segmentation and animal tracking. Zhang et al. \cite{zhang2023uvosam} proposed a novel unsupervised video object segmentation model,  UVOSAM, which does not require masks. This has opened up new possibilities in the field. Dutta et al. \cite{dutta2023robust} developed a deep learning workflow for animal tracking, which is particularly relevant for ecological studies. Javed et al. \cite{javed2022visual} provided a comprehensive survey of visual object tracking techniques, which has been instrumental in shaping our understanding of the current landscape. Cao et al. \cite{cao2022track} proposed a method of dense spatio-temporal position encoding to improve tracking accuracy. Proencca et al. \cite{proencca2023trade} introduced TRADE a method that utilises 3D trajectory and ground depth estimates for UAVs. Jahanbakht et al. \cite{jahanbakht2022distributed} explored distributed deep learning and energy-efficient image processing for fish segmentation. Finally, Zhang et al. \cite{zhang10msgnet} developed MSGNet, which uses multiple sources of information to improve the precision of fish segmentation. This approach has been influential in our own methodology.

% In the following subsection, we briefly review the research domains most relevant to our work.

\textbf{Video object segmentation}
is a task that is used to locate and segment each target object \cite{Yao2020, Khoreva2019vos, Maninis2019vos}. The target object to be segmented can be either a class of interest in the videos or moving objects of interest.
% is the task of separating all foreground objects from the background regions in the video \cite{}. 
Object segmentation is generally categorized into two categories: segmentation with instance-level semantics and segmentation without instance-level semantics, which is the main objective of this paper.
Therefore, this study focusses on generating labels without human intervention.
Some segmentation methods for moving objects have been developed by using background subtraction techniques \cite{Bouwmans2019bg, Kalsotra2019bg, Garcia2020bg}. 
Several of these approaches are based on the assumption that the scene is locally constant \cite{pan2021bg, Maddalena2018bg}. This means that the background in one frame is assumed to be similar to the background in the next frame or only a few pixels away. In order to use this assumption, they estimate the local background and threshold it according to the similarity threshold to identify foreground regions. 
However, this method is known to be sensitive to illumination changes and may even lose all detail within the image due to overestimating the local background. Another approach to segmentation uses the detection of optical flow to define motion boundaries \cite{lu2021of, Anthwal2019of, cheng2017of, ding2020of}. 

\textbf{Optical flow}
predicts the relative motion of objects in two consecutive frames of a video \cite{cheng2017of, ding2020of}.
It gives a dense correspondence between frames, but at the cost of being limited to rigid objects, and computation entangled. Additionally, optical flow can only work within scenes where the movement of the camera is significantly lower than the movement of the object \cite{lu2021of, Anthwal2019of, Garcia2014of}.
This can be seen as a limitation, as the background subtraction method can be used in a wider range of applications. However, the key element of optical flow is that it can also be used for background subtraction \cite{chraa2018of, Kushwaha2020of}. 
By tracking the movement of the pixels between frames, we can determine the background. If a pixel that is part of the background does not match the static background within a given threshold, then that pixel is determined to be an instance of an object. 

Another segmentation approach is based on the detection of visual motion. It is based on the fact that moving objects in the scene induce consistent changes to the flow of pixels in a region \cite{Anthwal2019of, Garcia2014of}.
However, due to substantial displacements or occlusions, their calculated optical flow may contain considerable inaccuracies \cite{sun2014of, chen2013of, brox2011of}.
In our method, we address these issues and enhance both estimated optical flow and object segmentation, simultaneously.

\textbf{Video object tracking}
is the task of assigning a consistent label for each individual object in the scene as it moves \cite{Guan2016vot, Ciaparrone2020vot}. This tracking is generally divided into multiple steps, including detection of the object of interest, tracking of the moving object in the scene, and then associating labels between frames. 
The tracking task, therefore, consists of identifying the bounding box of the object over several video frames and, at the same time, updating the location of the object in the image \cite{Gomez2022vot, qiu2020vot}. This can be done based on a similarity metric between different frames \cite{kang2019sm, Dadgar2021sm}. The idea of such a metric is to find the closest objects in the frame with an overlapping bounding box. This can be performed at either the pixel level or at the region level. 
The major drawback of this method is the computation time \cite{bag2019sm, zhu2021sm} that is needed to compute all the similarities between all the different frames. 
On the other hand, if the computational resources are available, this method has been proven to be useful when tracking fast-moving objects and when the objects are not occluded in the frame \cite{zhu2020mov, Chapel2020mov}.
In our method, we produce the rotating 2D object bounding box from each instance mask of the object over several video frames.

In contrast to our work, Yang \textit{et al.} \cite{kai2022siam} uses a Siamese network with an anchor-free tracker for general object tracking, simplifying the tracking algorithm by avoiding the anchor box design that predicts the tracking target with a pair of corners (top-left and bottom-right corners). 
While both our work and SiamCorners \cite{kai2022siam}  utilize deep learning techniques for object tracking, there are key differences. Our approach specifically addresses the challenges of fish segmentation and tracking in underwater videos, leveraging optical flow and background subtraction to generate pseudo-ground truth labels. In contrast, SiamCorners simplifies the tracking algorithm by avoiding anchor box design but does not specifically address the unique challenges of fish segmentation and tracking in underwater videos. We believe these distinctions highlight the novelty and significance of our work in this specific domain.

\textbf{Supervised And Unsupervised Learning.}
Supervised learning has been used to build object detection \cite{zhu2020od, Jiao2019od, Zhao2019od}, video object segmentation \cite{Khoreva2019vos, Maninis2019vos} and video object tracking \cite{Ciaparrone2020vot, Gomez2022vot}. 
These methods require extensive human annotation and therefore are not suitable for video annotation in the wild. To reduce the labelling costs of data, unsupervised learning has emerged as a powerful technique for the learning of video data. In the traditional image domain, unsupervised methods are expected to outperform their supervised counterparts \cite{Jiang2020, Wang2020c, Saleh2020, Konovalov2019, Konovalov2018} due to their potential to train data without labels. 

The idea behind many of the unsupervised \ac{DNN} models is to learn a feature representation from unlabelled data \cite{Zhou2020fr, Peng2020fr, xie2020fr}. Then, a \ac{DNN} model can be applied to the learned feature representation to produce the output. 
% The unsupervised \ac{DNN} models can provide a large-scale data structure with limited annotation costs, without the need to manually label videos. 
For example, in the domain of video segmentation \cite{Garcia2020a,Chang2021,Alshdaifat2020,Laradji2021}, \acp{DNN} have been used to learn a representation from the difference between a pair of unlabelled videos \cite{Jabri2020,Araslanov2021,Wang2020cor} and from warped frames \cite{liu2022warp}. 

In this work, we focus on unsupervised learning. Our proposed method will generate labels  referred to as pseudo-labels to train a multi-task supervised \ac{DNN} for video object segmentation and video object tracking.
\begin{figure*}[!t]
\centering
\includegraphics[width=0.88\textwidth]{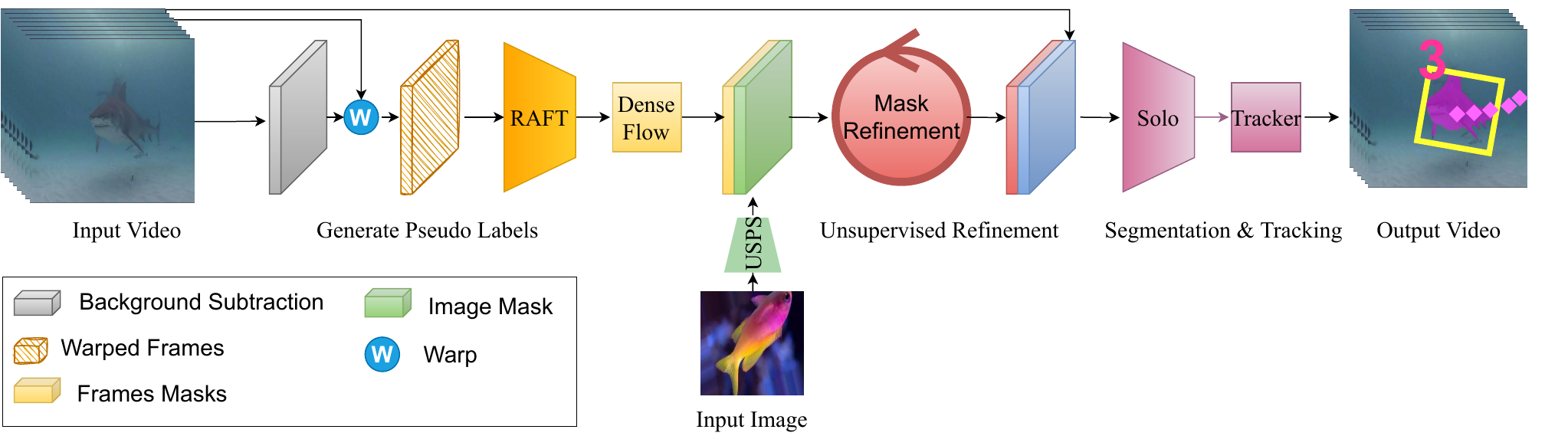}
\caption{Our proposed framework consists of three main components: generate pseudo-labels, unsupervised pseudo-labels refinement, and segmentation network. The proposed segmentation model trains with the generated pseudo-labels, which are refined with self-supervised training. Please refer to \cref{secfram} for details.}
\label{fig:1}
\end{figure*}
% %%%%%%%%%%%%%%%%%%%%%%%%%%%%%%%%%%%%%%%%%%%%%%%%%%%%%%%%%%%%%%%%

%%%%%%%%%%%%%%%%%%%%%%%%%%%%%%%%%%%%%%%%%%%%%%%%%%%%%%%%%%%%%%%%
\section{Framework} \label{secfram}
% The architecture of the proposed model is illustrated in 
The overall framework can be divided into three stages as shown in \cref{fig:1}.
% Our Framework is made of three main stages.
The first stage is to generate pseudo-labels using background subtraction and optical flow for both videos and still images.
The second stage is to train a self-supervised model to refine the pseudo-labels using their spatial structure.
In the last stage, the refinement of the video and still-image versions are applied jointly to train the segmentation network and to predict the final label. 
The segmentation network's training behaviour closely matches supervised training because we employ improved pseudo-labels. As a result, the network's training process is more reliable than that of current unsupervised learning techniques \cite{Saleh2022uvos, Jabri2020,Araslanov2021,Wang2020cor}.
In the following subsections, we describe the details of these three components and the corresponding loss functions. 

\subsection{Background Subtraction} \label{secbck}
As a first step to generating pseudo labels, background subtraction is performed on the video frames. A clean background image is estimated for every video sequence by computing the median of the first 10 frames of the video sequence along the first axis.  This is to average out any distracting elements that come in front of the clean background. Then, each video frame is subtracted from the clean background to create the mask sequence. After subtracting, all foreground pixels take on a value of 1, and pixels belonging to any background region have 0 values using Adaptive Gaussian Thresholding \cite{Golilarz2019gaussn}.

Adaptive Gaussian Thresholding is used instead of one global value as a threshold because it sets a pixel's threshold based on a local region surrounding it. As a consequence, we obtain various thresholds for various areas inside the same image, which produces better results for images with varying illumination.

This background subtraction step is crucial in eliminating any stationary elements or shadows from the video sequences that might disturb the next step, optical flow.

\subsection{Optical Flow} \label{secflow}
The next step in pseudo-label generation is to calculate the optical flow using \ac{RAFT} \cite{teed2021raft}.
However, optical flow is frequently inaccurate at object boundaries, so we want our segmentation to be accurate exactly at these borders.
Therefore, we consider video segmentation from background subtraction and optical flow estimation simultaneously. Using pixel level and temporal information sources, the segmentation algorithm is improved by removing artifacts induced by background subtraction and optical flow.
We demonstrate how both levels work in concert to preserve object boundaries and temporal coherence throughout the video.
The key is that we need to remove motion blurs while preserving the motion of the fish boundaries. 

To achieve the pseudo labels, we first deconstruct a pair of video frames, $x_t$ and $x_{t+1}$, and estimate a mask $m_t$ and $m_{t+1}$ with the background subtraction method as described in section \cref{secbck}. 
Segmented masks $m_t$, $m_{t+1}$ are used to synthesise frames $\hat{x_t}$ and $\hat{x_{t+1}}$ by warping $x_t$ and $x_{t+1}$ with $m_t$, $m_{t+1}$, respectively.
The optical flow \cite{teed2021raft} takes two frames $\hat{x_t}$ and $\hat{x_{t+1}}$, and produces a motion vector $\hat{v}$ between them.
This motion vector is used to compute the magnitude and angle of the motion. 
Specifically, pixels with a motion vector $\hat{v}$ outside $m_t$ (and $m_{t+1}$) are assigned the value of the background, and pixels with a motion vector $\hat{v}$ inside $m_t$ (and $m_{t+1}$) are reassigned the object. 
We denote the reassigned images as $\hat{x_t^*}$ and $\hat{x_{t+1}^*}$ and use them as input for our segmentation step, as shown in the top panel of  \cref{fig:1}. 

\begin{figure*}[ht]
\centering
\includegraphics[width=0.88\textwidth]{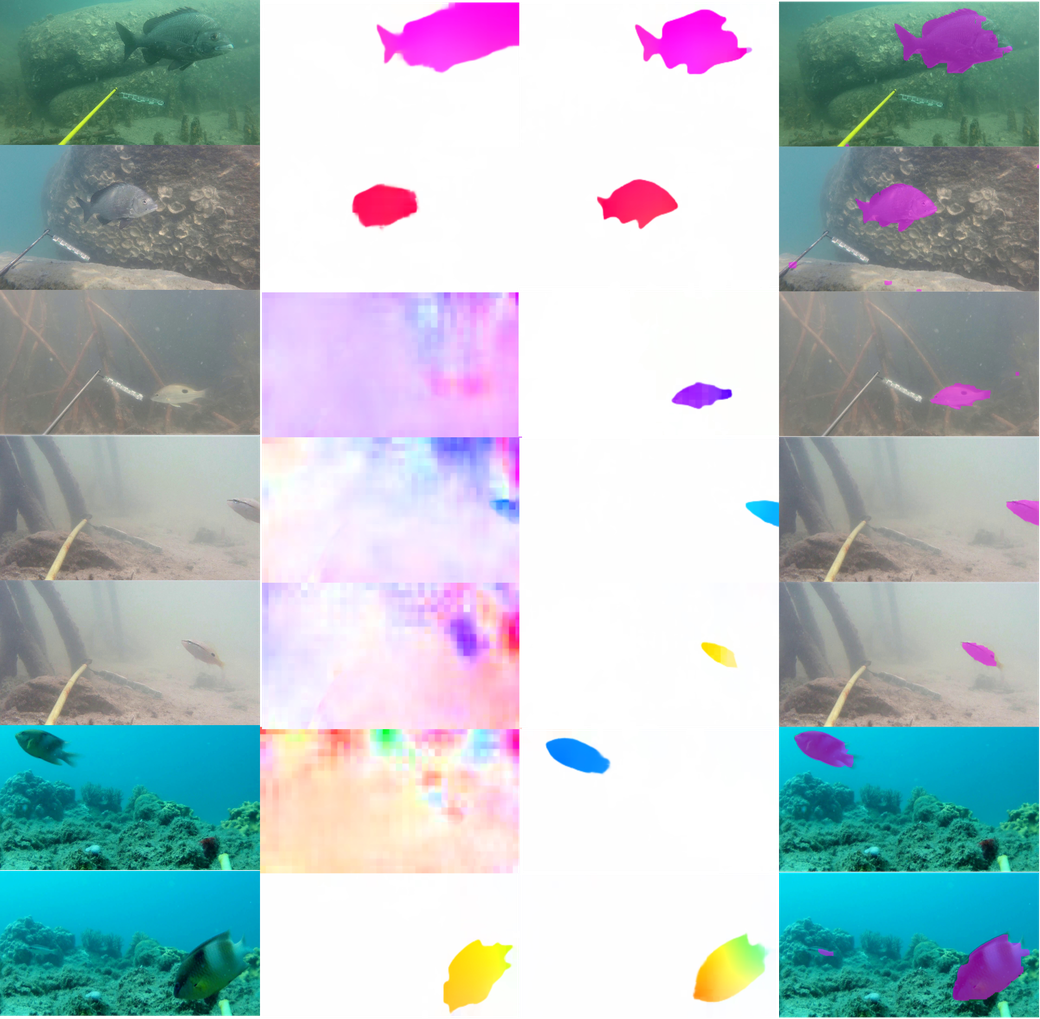}
\caption{Sample optical flow results for Seagrass \cite{Ditria2021a}. From left, the original image, optical flow without background subtraction, optical flow with background subtraction, mask overlay.}
\label{fig:3}
\end{figure*}

\begin{figure*}[ht]
\centering
\includegraphics[width=0.88\textwidth]{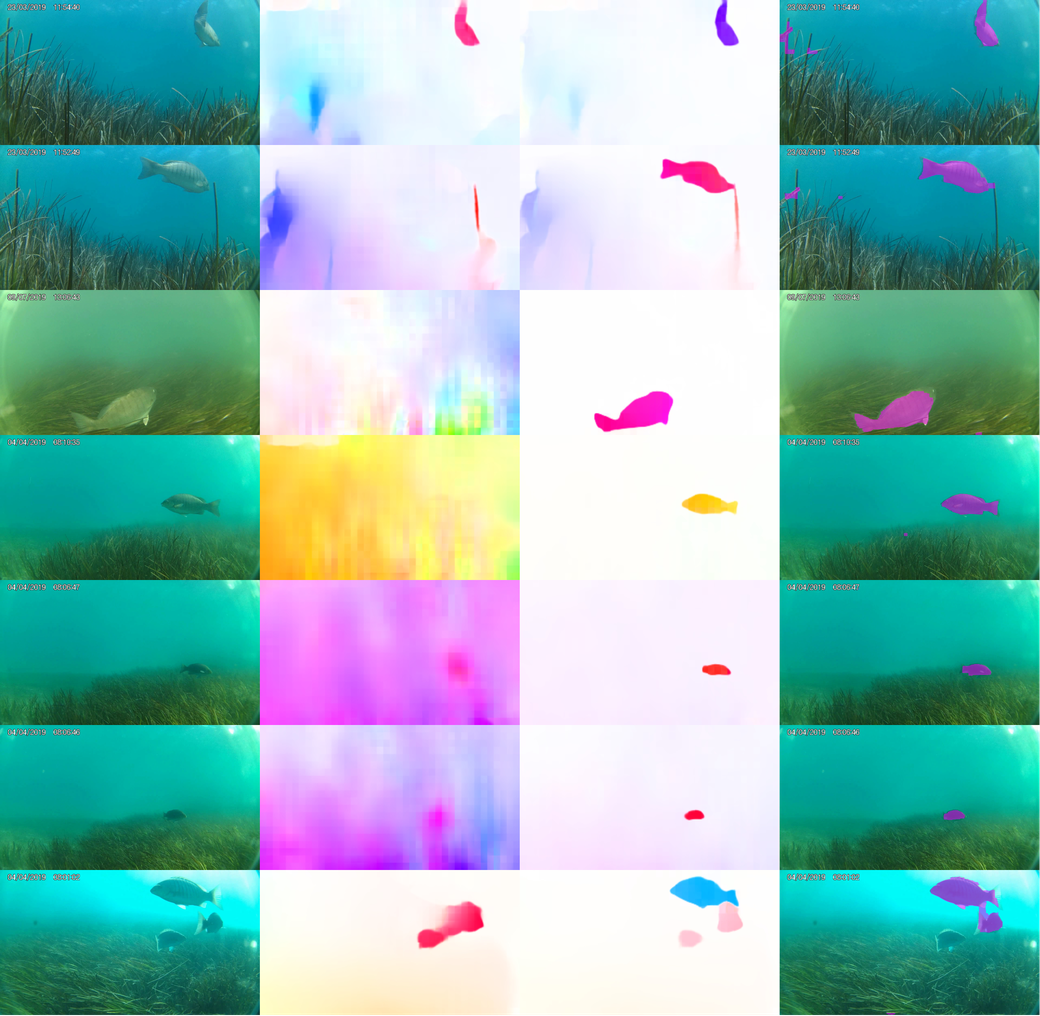}
\caption{Sample optical flow results for DeepFish \cite{Saleh2020a}. From left, the original image, optical flow without background subtraction, optical flow with background subtraction, mask overlay.}
\label{fig:4}
\end{figure*}

\begin{figure}[ht]
\centering
\includegraphics[width=0.48\textwidth]{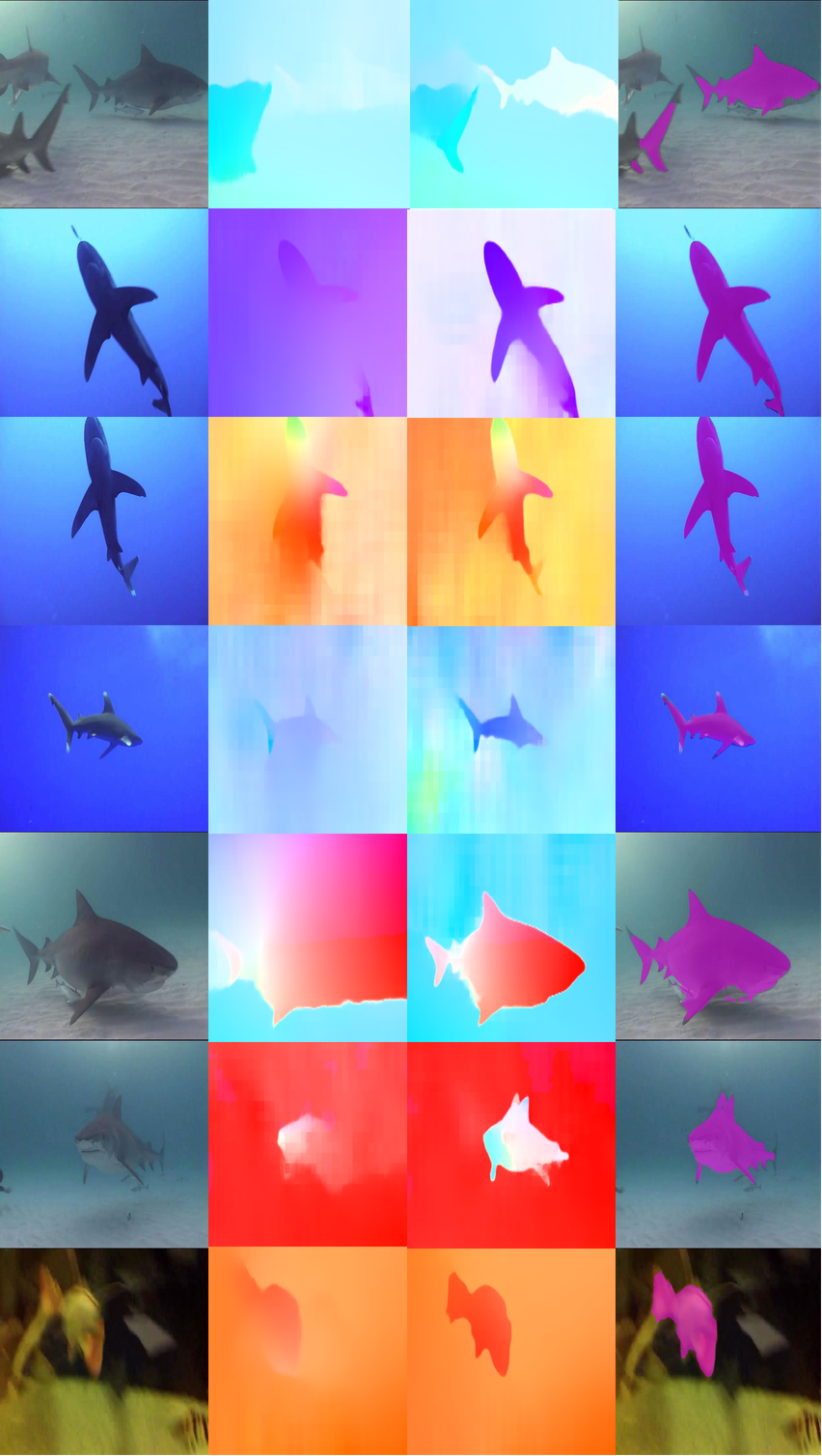}
\caption{Sample optical flow results for YouTube-VOS \cite{Xu2018b}. From left, the original image, optical flow without background subtraction, optical flow with background subtraction, mask overlay.}
\label{fig:5}
\end{figure}

We show the optical flow results for the three video datasets with and without background subtraction of frames $x_t$ and $x_{t+1}$ in  \cref{fig:3}, \cref{fig:4}, and \cref{fig:5}. A mask $m_{t+1}$ that better distinguishes the background from the foreground from the optical flow step is then refined with our proposed unsupervised refinement method in the next section.
A sample optical flow comparison video before and after background subtraction is available at  \href{https://youtu.be/8LOKsVSiY9U}{https://youtu.be/8LOKsVSiY9U}.

% We show the background subtraction results of frames $x_t$ and $x_{t+1}$ on the right of \cref{fig:3}. 
% The blue area is the background that is not segmented. In the right image of Fig.~\ref{figflow}, we show the reconstructed $\hat{x_t^*}$ and $\hat{x_{t+1}^*}$. The white background is replaced by the pixel from the next frame. A mask $m_{t+1}$ that can better distinguish between the background and foreground regions is obtained as shown in Fig.~\ref{figflow}. Our method produces the reconstruction $\hat{x_{t+1}^*}$ based on $\hat{x_t^*}$. 

% https://codeyarns.com/tech/2014-02-26-carttopolar-in-opencv.html#gsc.tab=0
%   https://www.programcreek.com/python/example/89301/cv2.cartToPolar
% These pseudo-labels are then refined by a self-supervised learning model.
% %%%%%%%%%%%%%%%%%%%%%%%%%%%%%%%%%%%%%%%%%%%%%%%%%%%%%%%%%%%%%%%%
\begin{figure*}[ht]
\centering
\includegraphics[width=0.98\textwidth]{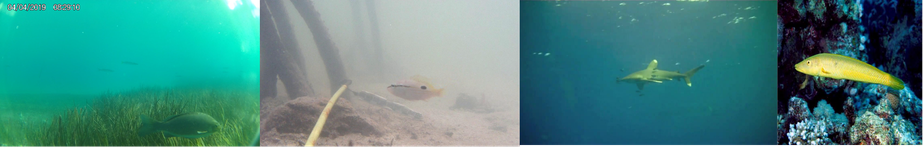}
\caption{Sample image from each of the four utilised datasets. From left: Seagrass \cite{Ditria2021a}, DeepFish \cite{Saleh2020a},  YouTube-VOS \cite{Xu2018b}, and Mediterranean Fish Species \cite{MediterraneanFish}}
\label{fig:2}
\end{figure*}
% %%%%%%%%%%%%%%%%%%%%%%%%%%%%%%%%%%%%%%%%%%%%%%%%%%%%%%%%%%%%%%%%

% %%%%%%%%%%%%%%%%%%%%%%%%%%%%%%%%%%%%%%%%%%%%%%%%%%%%%%%%%%%%%%%%
% %%%%%%%%%%%%%%%%%%%%%%%%%%%%%%%%%%%%%%%%%%%%%%%%%%%%%%%%%%%%%%%%

% %%%%%%%%%%%%%%%%%%%%%%%%%%%%%%%%%%%%%%%%%%%%%%%%%%%%%%%%%%%%%%%%
% %%%%%%%%%%%%%%%%%%%%%%%%%%%%%%%%%%%%%%%%%%%%%%%%%%%%%%%%%%%%%%%%

% %%%%%%%%%%%%%%%%%%%%%%%%%%%%%%%%%%%%%%%%%%%%%%%%%%%%%%%%%%%%%%%%
% %%%%%%%%%%%%%%%%%%%%%%%%%%%%%%%%%%%%%%%%%%%%%%%%%%%%%%%%%%%%%%%%

% %%%%%%%%%%%%%%%%%%%%%%%%%%%%%%%%%%%%%%%%%%%%%%%%%%%%%%%%%%%%%%%%
\begin{figure*}[ht]
\centering
\includegraphics[width=0.98\textwidth]{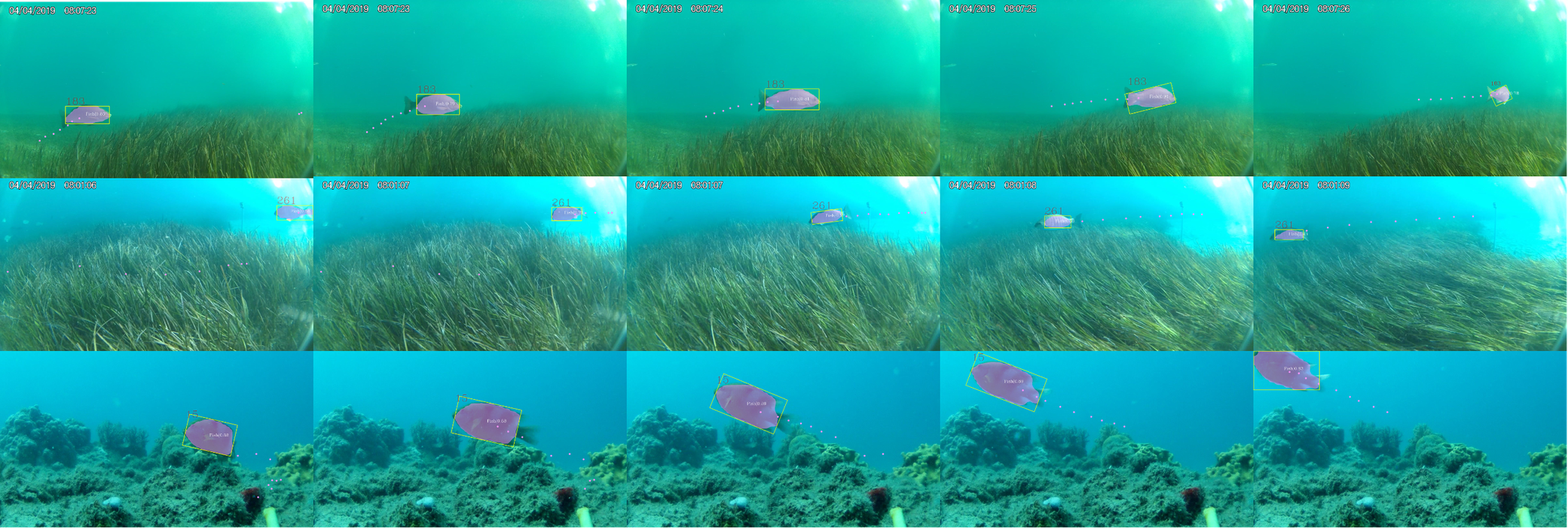}
\caption{Sample fish trajectory results. Zoom in for a better view. See also a short video of fish trajectory results at \href{https://youtu.be/Z5G7YBoL3eM}{https://youtu.be/Z5G7YBoL3eM}.}
\label{fig:9}
\end{figure*}
% %%%%%%%%%%%%%%%%%%%%%%%%%%%%%%%%%%%%%%%%%%%%%%%%%%%%%%%%%%%%%%%%

\subsection{Unsupervised Refinement} \label{secrefine}
The second stage in our method is cumulative pseudo-label refinement through unsupervised historical moving averages (MVA) \cite{Nguyen2019usps} using DeepLabv3 \cite{chen2018deeplab} network for semantic segmentation and Conditional Random Fields (CRF) \citep{krahenbuhl2011efficient} by minimising the F-score until the MVA predictions reach a stable state. The CRF can "sharpen" initial location predictions to make them more accurate and consistent with edges and parts of the source image that have a constant colour.

Given the pseudo-labels of the previous step, we train the network for $50$ epochs. The number of epochs is low to avoid a significant over-fitting of the network to the noisy pseudo-labels.
Then, the network is reinitialized with trained weight to predict a new set of pseudo-labels to train on again. 

Let $D$ be the set of training examples and $M$ be the network model. By $M(x, p)$ we denote the mask prediction of model $M$ on the pixel $p$ of the image $x \in D$.
During this stage, a historical moving average (MVA) from the last training stage is composed as follows:
\begin{equation*}
  \text{MVA}(x, p, k)=
    (1-\alpha) * CRF(M(x, p)) + \alpha * \text{MVA}(x, p, k-1),
\end{equation*}
where  $M(x, p)$ is the network mask prediction,  $k$ is the epoch number, $\alpha$ is a positive real factor, and $CRF$ is the Conditional Random Fields (CRF) \citep{krahenbuhl2011efficient}.

We use  $L_\beta= 1 - F_\beta$  as an \emph{image-level loss} function w.r.t. each training example $x$.
 F-score ($F_\beta$ ) is the harmonic mean of precision and recall of the prediction output of pixel $p$ on image $x$ w.r.t. the pseudo-labels, which use a positive real factor $\beta$ as follows:

\begin{equation*}
    F_\beta=\left(1+\beta^2\right)\frac{\text{precision} \cdot \text{recall}}{\beta^2~ \text{precision} + \text{recall}}.
\end{equation*}

The network is retrained until the MVA reaches a stable state, as shown in the middle panel of \cref{fig:1}.
By doing so, the quality of pseudo-labels is improved over time.

\subsection{Segmenting Objects by Locations} \label{secsolo}
Our last stage is training a supervised segmentation model using the refined pseudo-labels from the previous stage.
The supervised model is based on Segmenting Objects by Locations (SoloV2) \cite{Wang2020SOLOv2}.
SoloV2 is an updated version of Solo \cite{Wang2020solo}, a previous method for instance segmentation. The idea is to dynamically segment objects by location. 

Given an image as input, the network generates the object mask, then the object mask generation is decoupled into a mask kernel prediction and mask feature learning. Furthermore, matrix non-maximum suppression (MNMS) is applied to reduce inference overhead.
Specifically, SoloV2 is composed of two modules: 
(1) Dynamic Instance Segmentation
and (2) Matrix non-maximum suppression (MNMS).
The dynamic instance segmentation scheme dynamically segments objects by location by learning the mask kernels and mask features separately. The mask kernels are predicted dynamically by the \ac{FCN} \cite{Shelhamer2017} when classifying the pixels into different location categories, then constructing a unified mask feature representation for instance-aware segmentation. 
The non-maximum suppression process is achieved by performing NMS with a parallel matrix operation in one shot to reduce inference overhead and suppress duplicate predictions. Compared to the widely adopted multi-class NMS \cite{Neubeck2006}, where the sequential and recursive operations result in non-negligible latency, the parallel non-maximum suppression with matrix operation can achieve similar performance with much lower latency. The parallel processing strategy performs MNMS inference on-the-fly and enables processing at a high frame rate (\textit{$34$ frames per second}). 
For more details, we refer the reader to  \cite{Wang2020SOLOv2}.

\subsection{Rotating Bounding-box} \label{secbbox}
From each instance mask that we predicted from the previous stage, we are able to produce the rotating 2D object bounding box.
The minimum bounding rectangle (MBR) technique is used to obtain a rotated bounding box from a binary mask of the object.
We used OpenCV \cite{OpenCv} to find the minimum area of a rotated rectangle. 
It takes the binary mask of the object as an input and returns a Box2D structure that contains the following information: (centre (x, y), (width, height), angle of rotation). 
The output of this step is used to track the objects as discussed in the following section.

\subsection{Online Tracking} \label{sectrck}
We used Simple Online and Real-Time Tracking (SORT) \cite{Bewley2016sort} as an online tracking framework that focuses on frame-to-frame prediction and association. 
The position and size of the bounding box are used only for both motion estimation and data association. 
Kalman filter \cite{Kalman1960} is used to handle motion estimation and the Hungarian method \cite{Kuhn1955} is used for data association.

Motion estimation is used to propagate a target’s identity into the next frame. 
The inter-frame displacements of each object are approximated with a linear constant velocity estimation.
The detected bounding box is used to update the target state where the Kalman filter \cite{Kalman1960} solves the velocity components.
The state of each target is estimated as:
$$x = [h, v, s, r, \hat{h} , \hat{v} , \hat{s}]^T,$$
where $h$ and $v$ represent the horizontal and vertical pixel location
of the centre of the target, while  $s$ and $r$ represent the scale and the aspect ratio of the target’s bounding box, respectively. Here, $\hat{h}, \hat{v}, \hat{s}$ are for the source.

Data association is assigning new detections to existing targets. Each target’s
bounding box is estimated by predicting its new location in the current frame.
The intersection-over-union ($IOU$) distance between each detection and each forecasted bounding box from the existing targets is used to calculate the assignment cost matrix. The assignment cost matrix is then resolved using the Hungarian technique \cite{Kuhn1955} to produce the fish trajectory as shown in \cref{fig:9}.

%%%%%%%%%%%%%%%%%%%%%%%%%%%%%%%%%%%%%%%%%%%%%%%%%%%%%%%%%%%%%%%%
\section{Method} \label{secmethod}
This section describes our method in detail. Our method is based on three main components: the pseudo labels generation, the unsupervised learning method to refine the generated pseudo labels and the \ac{DNN} for fish tracking and segmentation. 
\cref{fig:1} shows the algorithm flow diagram for the fish tracking and segmentation framework.

\subsection{Datasets} \label{secdata}
We performed experiments using four publicly available datasets, i.e.  Seagrass \cite{Ditria2021a}, DeepFish \cite{Saleh2020a},  YouTube-VOS \cite{Xu2018b}, and Mediterranean Fish Species \cite{MediterraneanFish}.
 \cref{fig:2} demonstrates a sample image from each dataset.

\textit{Seagrass} \cite{Ditria2021a} 
is comprised of annotated footage of \textit{Girella tricuspidata} in two estuary systems in south-east Queensland, Australia. 
The raw data was obtained using submerged action cameras (HD $1080p$).
The dataset includes $4280$ video frames and $9429$ annotations.
Each annotation includes segmentation masks that outline the species as a polygon.

\textit{DeepFish} \cite{Saleh2020a} 
consists of a large number of videos collected from 20 different habitats in remote coastal marine environments of tropical Australia. 
The video clips were captured in full HD resolution ($1920 \times 1080$ pixels) using a digital camera. In total, the number of video frames taken is about $40k$. 

\textit{YouTube-VOS} \cite{Xu2018b} 
is a video object segmentation dataset that contains $4453$ YouTube video clips and 94 object categories. The videos have pixel-level ground truth annotations for every $5th$ frame ($6 fps$). 
For a fair comparison, we extracted only the videos that contained \texttt{fish}, which include $130$ video clips and $4349$ video frames in total.

\textit{Mediterranean Fish Species} \cite{MediterraneanFish} 
consists of a large number of images collected from 20 different Mediterranean fish species. In total, the number of images is about $40k$. 
The dataset was split into two subfolders, training and test sets. 
The training set contains $34k$  and the test set contains $6k$ images.
The image resolution ranges between ($220\times 210$ pixels) and ($1920 \times 1080$ pixels). The original images are stored in an RGB file format in subfolders as a class label.

We train our feature extractor on all of the four datasets and evaluate it on the video datasets only, Seagrass \cite{Ditria2021a}, DeepFish \cite{Saleh2020a},  and YouTube-VOS \cite{Xu2018b}.

\subsection{Pseudo Labeling} \label{secpsdo}
To train our supervised model, which was explained in \cref{secsolo}, we first generate pseudo labels for the image dataset, Mediterranean Fish Species \cite{MediterraneanFish}  and the video datasets, Seagrass \cite{Ditria2021a}, DeepFish \cite{Saleh2020a},  and YouTube-VOS \cite{Xu2018b}.

\subsubsection{Image Dataset}
Since our image dataset \cite{MediterraneanFish} is curated from static images of different fish species, our framework discussed in \cref{secfram} was not applicable to this dataset. Therefore, we used DeepUSPS \cite{Nguyen2019usps} as an unsupervised saliency prediction network for a pseudo-labels generation.
DeepUSPS is trained on the unlabelled MSRA-B dataset \cite{Liu2010mrsa} for predicting salient objects.
And it is an unsupervised learning method that produces pseudo labels with high intra-class variations, which is useful for the training of the supervised model.

However, DeepUSPS is only good in pseudo prediction for a single object in the image that is not disturbed by additional intricate details, which is not ideal for the more challenging video datasets \cite{Ditria2021a, Saleh2020a, Xu2018b}.

\subsubsection{Video Datasets}
Unlike our image dataset, our video datasets contain multiple objects in a single frame as well as across multiple frames. Therefore, we adapted our pseudo-label generation framework discussed in \cref{secfram}  that is capable of predicting multiple salient objects in the same video clip and handling the case of a cluttered background. This pseudo-label generation framework aims to tackle the issue of single-image datasets by generating more pseudo labels with intra-class variations in image space.

The Pseudo-label generation framework consists of three steps:
\begin{enumerate}
    \item Obtain salient objects by performing background subtraction using Adaptive Gaussian Thresholding \cite{Golilarz2019gaussn}, as explained in \cref{secbck}.
    \item Enhance the obtained salient object boundaries from the previous step with optical flow using RAFT \cite{teed2021raft}, as explained in \cref{secflow}.
    \item Apply cumulative pseudo-label refining via unsupervised historical moving averages (MVA) \cite{Nguyen2019usps}, as explained in \cref{secrefine}. 
\end{enumerate}
In this way, we can get pseudo labels for video datasets, Seagrass \cite{Ditria2021a}, DeepFish \cite{Saleh2020a}, and YouTube-VOS \cite{Xu2018b}, which are used to train the supervised model.

\subsection{Model training} \label{sectrain}

Our models were trained with an input resolution of $256 \times 256$ pixels.
We scale the lowest side of the video frames to $256$ and then extract random crops of size $256 \times 256$. We sample two video sets, $B = 2$ (of size $T = 5$ frames), therefore, $B \times T = 2 \times 5 = 10$ frames are used per forward pass.

We found that for this problem set, a learning rate of $1 \times 10^{-3}$ works the best. It took around 300 epochs for all models to train on this problem. Our networks were trained on a  Linux host with a single NVidia GeForce RTX 2080 Ti GPU with $11$ GB of memory, using Pytorch framework \cite{Paszke2019}.  We used stochastic gradient descent (SGD) optimiser \cite{Kingma2014Adam:Optimization}
with an initial learning rate of $0.01$, which is then divided by $10$ at
$27th$ and again at $33th$ epoch.
We use light augmentation (resizing, grayscale). Following \cite{chen2019tmask, Wang2020SOLOv2}, a scale jitter is used, where the shorter image side is randomly sampled from 640 to 800 pixels.

We applied the same hyperparameter configuration for all of the models.
However, the optimum model configuration will depend on the application, hence, these results are not intended to represent a complete search of model configurations. 
% Our training loop  is shown in algorithm \ref{code2}.

\subsection{Inference} \label{secinfr}
During tracking, we extract frames from the input video, forward each frame through the network, and obtain the fish category score from the classification branch. 
Initially, to filter out predictions with low confidence, we use a threshold of $0.1$ and perform convolution on the mask feature using corresponding predicted mask kernels.
Then, after applying a per-pixel sigmoid, we binarise the output of the mask branch at the threshold of $0.5$.
The final step is the matrix NMS, which fits the output mask with the Min-max box.

Our model operates online without any adaptation to the video sequence. On a single NVidia GeForce RTX 2080 Ti GPU, we measured an average speed of $34$ frames per second.

%%%%%%%%%%%%%%%%%%%%%%%%%%%%%%%%%%%%%%%%%%%%%%%%%%%%%%%%%%%%%%%%
\section{Experiments} \label{secresult}
We report experimental results for our model's trained representation on 50\% of the DeepFish, Seagrass, YouTube-VOS datasets and the train set of the Mediterranean Fish Species dataset. We then evaluated it in the other 50\% of the first three datasets. 
We provide quantitative and qualitative results that demonstrate our model's generalisation capabilities to a range of different underwater habitats.

\subsection{Results} \label{secmetrc}
We summarize our main results on Seagrass \cite{Ditria2021a}, DeepFish \cite{Saleh2020a}  and YouTube-VOS \cite{Xu2018b} datasets in \cref{tab:result}.
The quantitative results for all datasets were obtained using the COCO dataset \cite{Lin2014} evaluation script.
% We used detection and segmentation evaluation metrics used by COCO dataset \cite{Lin2014}.
The average precision ($AP$), the average recall ($AR$), and  Intersection over Union ($IoU$) were measured for the predicted bounding boxes and segmentation masks in the output images obtained from the trained SoloV2 \cite{Wang2020SOLOv2}, as explained in \cref{secsolo} in detail.

The Average Precision ($AP$) and Average Recall ($AR$) metrics provide a comprehensive view of the model's performance. The values $AP^{.50}$ and $AP^{.75}$ indicate that the model has a high precision rate when the Intersection Over Union (IoU) thresholds are 0.5 and 0.75, respectively. This means that the model is able to accurately predict the bounding boxes and segmentation masks for a majority of the objects in the images.
The values $AP^{M}$ and $AR^{M}$ show that the model maintains its precision and recall across a range of IoU thresholds, indicating its robustness to variations in object size and shape.
The $AP^{L}$ and $AR^{L}$ values specifically measure the model's performance on large objects. These metrics are particularly important in our case, as they reflect the model's ability to accurately segment and track larger fish species.

The results across different datasets demonstrate that our model is capable of generalizing well to unseen videos in other environments. This is a significant achievement, as it suggests that our approach could be applied to a wide range of underwater video data.

To the best of our knowledge, no prior research has reported detection and segmentation evaluation for these datasets. To compare our proposed unsupervised method to a supervised approach, we present the results of SoloV2 \cite{Wang2020SOLOv2} in the three data sets in \cref{tab:result2}. This Table displays the results of a fully supervised model with the original labels, not our generated pseudo-label.
 
In both tables, \cref{tab:result} and \cref{tab:result2}, higher values are better because they indicate that the model's predictions are more accurate.
From these tables, we can see that both unsupervised and supervised methods perform well across all three datasets, with some variation in performance depending on the specific dataset and whether detection or segmentation was being evaluated.

For example, in Table 1 (unsupervised method), we can see that the model performs best on the DeepFish dataset in terms of segmentation ($AP^{M} = 31.2$, $AR^{M} = 56.6$), but struggles more with detection on this dataset ($AP^{M} = 11.7$, $AR^{M} = 34.5$).

In contrast, in Table 2 (supervised method), we can see that although performance generally improves across all metrics compared to the unsupervised method, there are still some challenges with certain datasets - for example, detection on the DeepFish dataset ($AP^{M} = 12.2$, $AR^{M} = 41.0$).

Our proposed unsupervised method has yielded close accuracy results to the original supervised SoloV2 \cite{Wang2020SOLOv2} in both detection and segmentation experiments, validating the effectiveness of our generative approach. Furthermore, our results suggest that the proposed model is not heavily impacted by different underwater habitats, with almost similar performance for DeepFish \cite{Saleh2020a} and Seagrass \cite{Ditria2021a} datasets. The latter is particularly challenging due to the difficulty of visually detecting the fish. In some cases, the proposed model is not as good as fully-supervised approaches. However, the primary objective of this study is the development of an unsupervised method for fish tracking and segmentation. We postulate that our proposed approach offers enhanced stability during training compared to other unsupervised methods without a dedicated pseudo-label generation step. This stability, coupled with the robust performance of our method across diverse datasets, underscores its potential for further refinement and application in this domain.

\begin{table}[ht]
\centering
\caption{Comparison of \textbf{*unsupervised*} detection and segmentation on Seagrass \cite{Ditria2021a}, DeepFish \cite{Saleh2020a}  and YouTube-VOS \cite{Xu2018b} datasets.}
\label{tab:result}
\resizebox{\linewidth}{!}{
\begin{tabular}{@{}lcccccc@{}}
\toprule
 Dataset  & $AP^{M}$ & $AP^{.50}$ & $AP^{.75}$ & $AP^{L}$ & $AR^{M}$ & $AR^{L}$   \\
\midrule
\multicolumn{6}{@{}l}{\textit{Evaluating Detection:}} \\
\cdashline{1-7}\\
Seagrass \cite{Ditria2021a}     & 22.1 & 72.5 & 13.7 & 38.2 & 61.4 & 61.3 \\
DeepFish \cite{Saleh2020a}      & 11.7 & 35.0 & 05.3 & 19.3 & 34.5 & 57.1 \\
YouTube-VOS \cite{Xu2018b}      & 23.6 & 43.2 & 18.4 & 26.9 & 46.1 & 57.5 \\
\midrule
\multicolumn{6}{@{}l}{\textit{Evaluating Segmentation:}} \\
\cdashline{1-7}\\
Seagrass \cite{Ditria2021a}     & 12.0 & 37.6 & 05.2 & 20.8 & 31.2 & 52.0 \\
DeepFish \cite{Saleh2020a}      & 31.2 & 75.0 & 24.4 & 43.8 & 56.6 & 59.4 \\
YouTube-VOS \cite{Xu2018b}      & 15.4 & 33.0 & 12.2 & 19.2 & 33.8 & 42.2 \\
\bottomrule 
\end{tabular}
}
\end{table}

\begin{figure}[!t]
\centering
\includegraphics[width=0.88\linewidth]{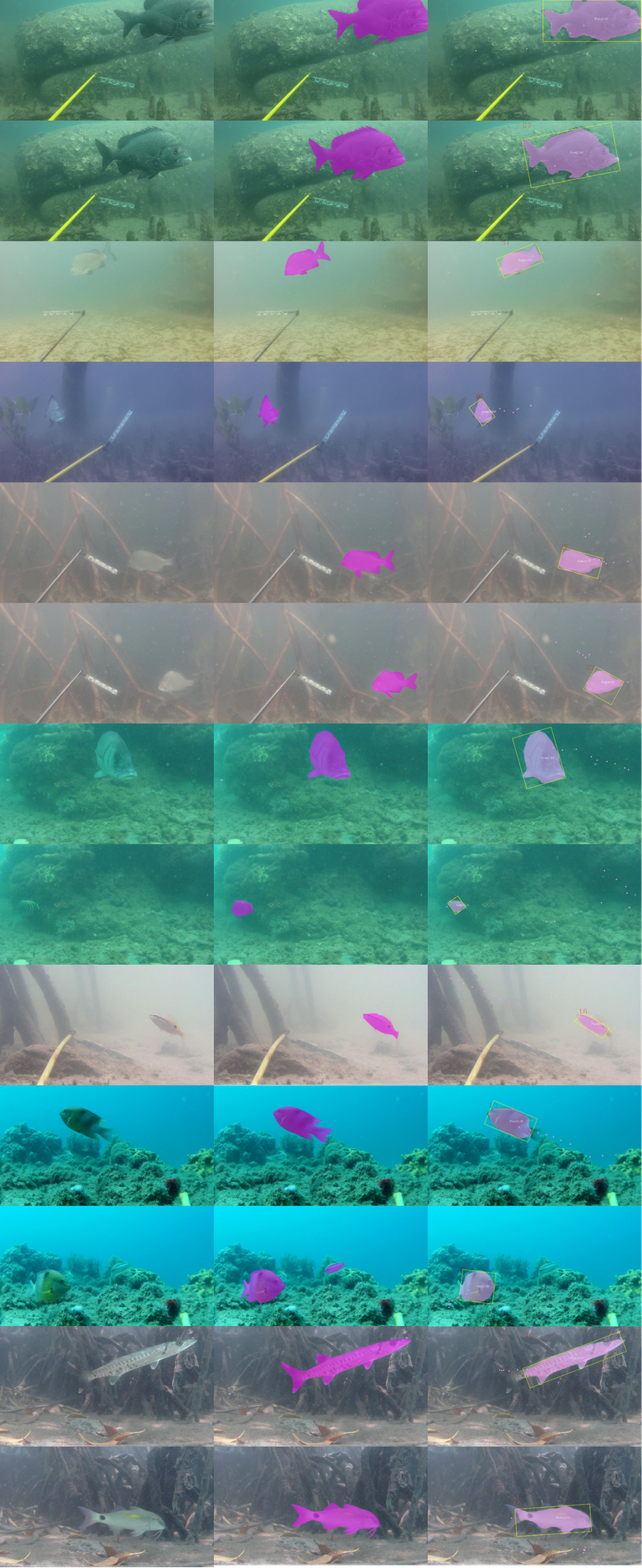}
\caption{Sample images from our model results for DeepFish \cite{Saleh2020a}, From left, the original image, the ground truth, the predicted image.}
\label{fig:6}
\end{figure}

\begin{figure}[ht]
\centering
\includegraphics[width=0.88\linewidth]{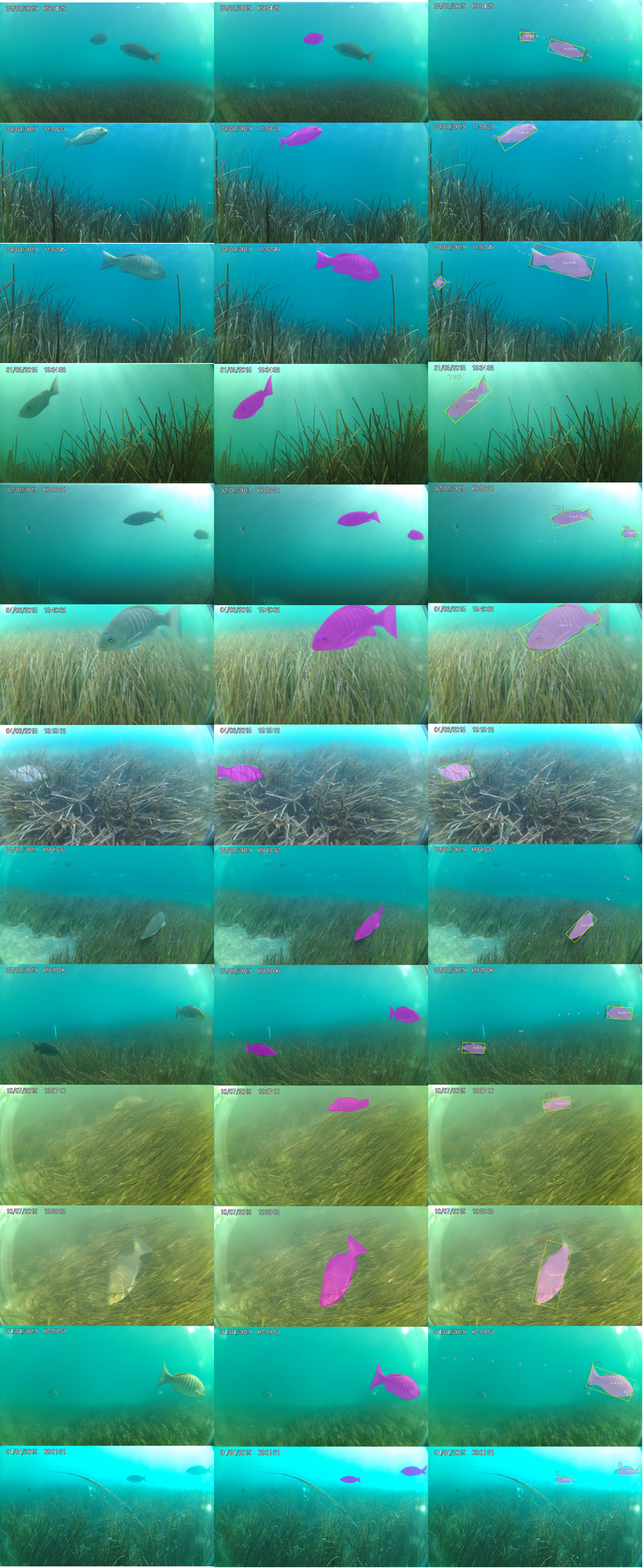}
\caption{Sample images from our model results for Seagrass \cite{Ditria2021a}, From left, the original image, the ground truth, the predicted image.}
\label{fig:7}
\end{figure}

\begin{figure}[ht]
\centering
\includegraphics[width=0.88\linewidth]{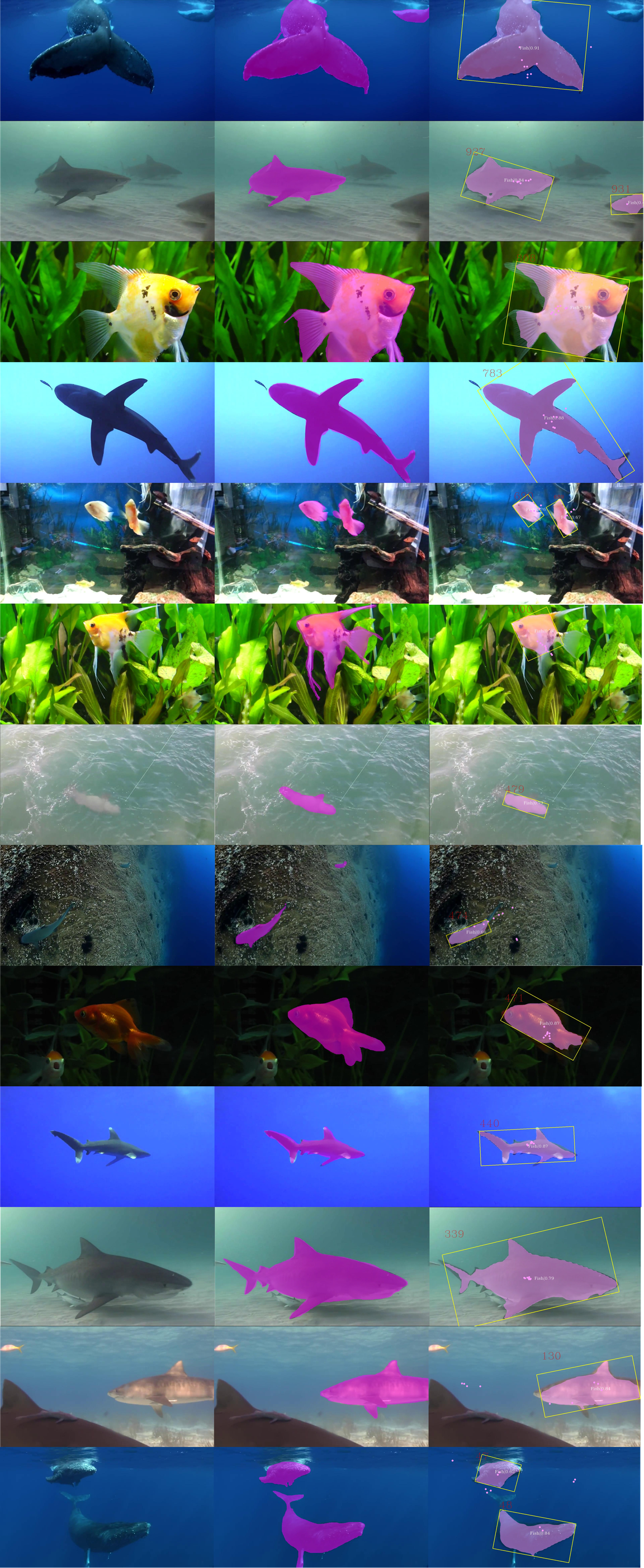}
\caption{Sample images from our model results for YouTube-VOS \cite{Xu2018b}, From left, the original image, the ground truth, the predicted image.}
\label{fig:8}
\end{figure}

\begin{figure}[ht]
\centering
\includegraphics[width=0.88\linewidth]{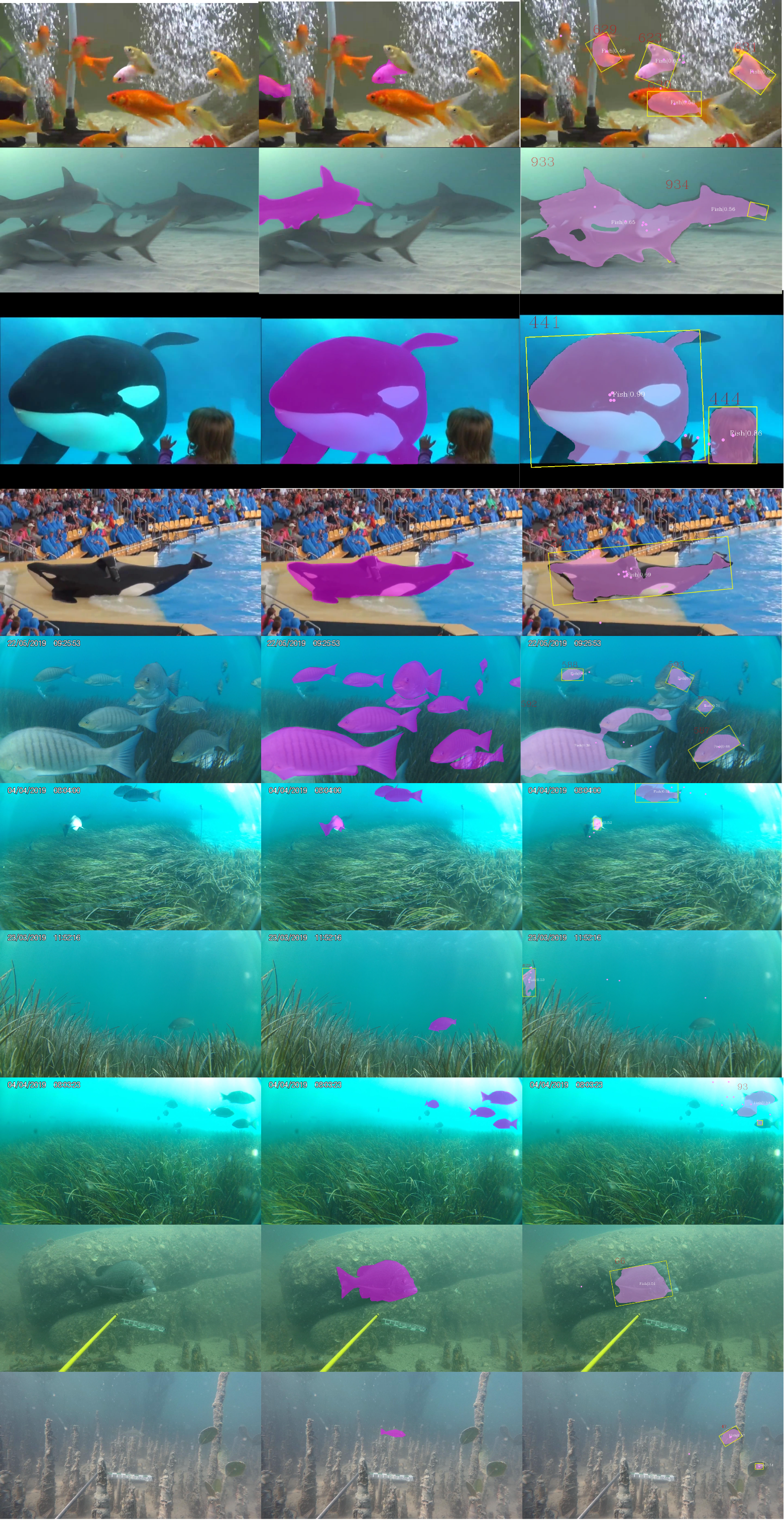}
\caption{
illustrate a sample of the failure cases of our model. From the left, the original image, the ground truth mask overlay, and the predicted image.
Images show instances where the model struggles with heavy occlusion, variability in fish size and shape, segmentation of foreground items, and influence of training videos. These scenarios highlight the limitations of our current approach and provide directions for future improvements.}
\label{fig:11}.
\end{figure}

The qualitative results of our algorithm for the DeepFish \cite{Saleh2020a}, Seagrass \cite{Ditria2021a} and YouTube-VOS \cite{Xu2018b} datasets are illustrated in  \cref{fig:6}, \cref{fig:7} and \cref{fig:8}, respectively.  Additional examples of failure cases are provided in \cref{fig:11}. 

Despite the challenges posed by fast movements and complex, crowded backgrounds, which often result in significant distortion, our algorithm produces favourable outcomes for the majority of images. This is particularly noteworthy for non-rigid objects.

For a more dynamic view of our model's predictions, you can watch a short video at this link \href{https://youtu.be/Z5G7YBoL3eM}{https://youtu.be/Z5G7YBoL3eM}. The video showcases the performance of our model in various scenarios, further demonstrating its effectiveness.

% Qualitative results of our algorithm for  DeepFish \cite{Saleh2020a}, Seagrass \cite{Ditria2021a} and YouTube-VOS \cite{Xu2018b} datasets are shown in \cref{fig:6}, \cref{fig:7} and \cref{fig:8}, respectively. 
% We also include additional examples of failure cases in \cref{fig:11}.
% Overall, especially for non-rigid objects, the proposed algorithm produces favourable outcomes in the majority of images. This is despite the fast movements or crowded and complicated backgrounds causing these images to frequently have significant distortion.
% See also a short video of our model's prediction at \href{https://youtu.be/Z5G7YBoL3eM}{https://youtu.be/Z5G7YBoL3eM}.

\begin{table}[ht]
\centering
\caption{Comparison of \textbf{*supervised*} detection and segmentation on Seagrass \cite{Ditria2021a}, DeepFish \cite{Saleh2020a}  and YouTube-VOS \cite{Xu2018b} datasets.}
\label{tab:result2}
\resizebox{\linewidth}{!}{
\begin{tabular}{@{}lcccccc@{}}
\toprule
 Dataset  & $AP^{M}$ & $AP^{.50}$ & $AP^{.75}$ & $AP^{L}$ & $AR^{M}$ & $AR^{L}$   \\
\midrule
\multicolumn{6}{@{}l}{\textit{Evaluating Detection:}} \\
\cdashline{1-7}\\
Seagrass \cite{Ditria2021a}     & 32.4 & 82.2 & 13.0 & 34.9 & 68.5 & 72.4 \\
DeepFish \cite{Saleh2020a}      & 12.2 & 41.8 & 04.3 & 20.9 & 41.0 & 68.0 \\
YouTube-VOS \cite{Xu2018b}      & 25.9 & 56.2 & 21.6 & 32.8 & 54.1 & 69.9 \\
\midrule
\multicolumn{6}{@{}l}{\textit{Evaluating Segmentation:}} \\
\cdashline{1-7}\\
Seagrass \cite{Ditria2021a}     & 18.0 & 56.4 & 07.8 & 31.2 & 36.8 & 68.0 \\
DeepFish \cite{Saleh2020a}      & 46.8 & 72.5 & 36.6 & 50.7 & 64.9 & 72.1 \\
YouTube-VOS \cite{Xu2018b}      & 23.1 & 49.5 & 18.3 & 28.8 & 40.7 & 53.3 \\
\bottomrule 
\end{tabular}
}
\end{table}

% \subsection{Qualitative Results} \label{secqlt}
%  \cref{fig:5} shows the qualitative results of our model using 

% %%%%%%%%%%%%%%%%%%%%%%%%%%%%%%%%%%%%%%%%%%%%%%%%%%%%%%%%%%%%%%%%
% %%%%%%%%%%%%%%%%%%%%%%%%%%%%%%%%%%%%%%%%%%%%%%%%%%%%%%%%%%%%%%%%

% %%%%%%%%%%%%%%%%%%%%%%%%%%%%%%%%%%%%%%%%%%%%%%%%%%%%%%%%%%%%%%%%
% %%%%%%%%%%%%%%%%%%%%%%%%%%%%%%%%%%%%%%%%%%%%%%%%%%%%%%%%%%%%%%%%

% %%%%%%%%%%%%%%%%%%%%%%%%%%%%%%%%%%%%%%%%%%%%%%%%%%%%%%%%%%%%%%%%
% %%%%%%%%%%%%%%%%%%%%%%%%%%%%%%%%%%%%%%%%%%%%%%%%%%%%%%%%%%%%%%%%

% %%%%%%%%%%%%%%%%%%%%%%%%%%%%%%%%%%%%%%%%%%%%%%%%%%%%%%%%%%%%%%%%
% %%%%%%%%%%%%%%%%%%%%%%%%%%%%%%%%%%%%%%%%%%%%%%%%%%%%%%%%%%%%%%%%

\subsection{Ablation Study} \label{secabl}
We performed an ablation study to demonstrate the proposed approach's effectiveness in generating pseudo-labels. Specifically, we analysed the contribution of the vital component in the proposed method, the optical flow with background subtraction (\cref{secflow}).
In addition, we evaluated the segmentation network training with refined pseudo-labels (\cref{secsolo}) for different epochs. 
The results reported in \cref{tab:abl2} are for unsupervised segmentation based on optical flow without background subtraction as a baseline.
And the results reported in \cref{tab:abl} are for the four epoch trials with the same random seeds, see \cref{sectrain} for the details.

The metrics used in the ablation study are as follows:
\begin{enumerate}
    \item $AP^{M}$ (Average Precision for Medium objects): This is the average precision for medium-sized objects. Precision is the ratio of correctly predicted positive observations to the total predicted positive observations. The higher the $AP^{M}$, the better the model is at predicting medium-sized objects correctly.

    \item $AP^{.50}$ and $AP^{.75}$: These are the average precision values at different Intersection over Union (IoU) thresholds. IoU is a measure of overlap between two bounding boxes. $AP^{.50}$ is the average precision when IoU is 0.50, and $AP^{.75}$ is the average precision when IoU is 0.75. Higher values indicate better precision at these IoU thresholds.

    \item $AP^{L}$ (Average Precision for Large Objects): This is the average precision for large-sized objects. Like $AP^{M}$, a higher $AP^{L}$ indicates that the model is better at predicting large-sized objects correctly.

    \item $AR^{M}$ (Average Recall for Medium Objects): This is the average recall for medium-sized objects. Recall is the ratio of correctly predicted positive observations to all actual positives. The higher the $AR^{M}$, the better the model is at identifying all actual medium-sized objects.

    \item $AR^{L}$ (Average Recall for Large Objects): This is the average recall for large-sized objects. Like $AR^{M}$, a higher $AR^{L}$ indicates that the model is better at identifying all actual large-sized objects.
\end{enumerate}
In all these metrics, higher values are better because they indicate that the model's predictions are more accurate.

It is apparent from the results that the segmentation accuracy of our proposed method has improved significantly when compared to that of the baseline method. 
We also note that the accuracy of the models also depends on the number of epochs used in the training. We observe from the results shown in \cref{tab:abl} that the segmentation accuracy decreases after  $100$ epochs.
The reason for this is the over-fitting of the network to the noisy pseudo-labels.
While the training losses for both the baseline and our model decreased, the segmentation accuracy for our model was still greater than that for the baseline.

\begin{table}[!t]
\centering
\caption{Comparison of \textbf{*unsupervised*} segmentation based on optical flow  \textit{without} background subtraction.}
\label{tab:abl2}
\resizebox{\linewidth}{!}{
\begin{tabular}{@{}lcccccc@{}}
\toprule
 Dataset  & $AP^{M}$ & $AP^{.50}$ & $AP^{.75}$ & $AP^{L}$ & $AR^{M}$ & $AR^{L}$   \\
\midrule
\multicolumn{6}{@{}l}{\textit{Evaluating Segmentation:}} \\
\cdashline{1-7}\\
Seagrass \cite{Ditria2021a}     & 05.0 & 23.8 & 03.1 & 14.7 & 19.7 & 29.5 \\
DeepFish \cite{Saleh2020a}      & 15.3 & 44.8 & 13.6 & 33.5 & 42.7 & 37.4 \\
YouTube-VOS \cite{Xu2018b}      & 07.2 & 23.8 & 07.4 & 11.9 & 26.1 & 33.0 \\
\bottomrule 
\end{tabular}
}
\end{table}

\begin{table}[!t]
\centering
\caption{Comparison of \textbf{*unsupervised*} segmentation for different epochs: 50,100,150,300}
\label{tab:abl}
\resizebox{\linewidth}{!}{
\begin{tabular}{@{}lcccccc@{}}
\toprule
Dataset  & $AP^{M}$ & $AP^{.50}$ & $AP^{.75}$ & $AP^{L}$ & $AR^{M}$ & $AR^{L}$   \\
\midrule
\multicolumn{6}{@{}l}{\textit{50 epochs:}} \\
\cdashline{1-7}\\
Seagrass \cite{Ditria2021a}     & 12.4 & 33.6 & 07.2 & 20.4 & 28.4 & 47.2 \\
DeepFish \cite{Saleh2020a}      & 32.0 & 68.6 & 30.8 & 34.8 & 53.6 & 56.2 \\
YouTube-VOS \cite{Xu2018b}      & 15.8 & 34.0 & 13.8 & 19.8 & 33.8 & 42.2 \\
\midrule
\multicolumn{6}{@{}l}{\textit{100 epochs:}} \\
\cdashline{1-7}\\
Seagrass \cite{Ditria2021a}     & 12.0 & 37.6 & 05.2 & 20.8 & 31.2 & 52.0 \\
DeepFish \cite{Saleh2020a}      & 31.2 & 75.0 & 24.4 & 43.8 & 56.6 & 59.4 \\
YouTube-VOS \cite{Xu2018b}      & 15.4 & 33.0 & 12.2 & 19.2 & 33.8 & 42.2 \\
\midrule
\multicolumn{6}{@{}l}{\textit{150 epochs:}} \\
\cdashline{1-7}\\
Seagrass \cite{Ditria2021a}     & 12.0 & 36.0 & 04.8 & 20.4 & 30.0 & 48.8 \\
DeepFish \cite{Saleh2020a}      & 30.4 & 69.8 & 23.2 & 32.4 & 54.2 & 56.8 \\
YouTube-VOS \cite{Xu2018b}      & 15.2 & 34.0 & 14.0 & 20.2 & 32.8 & 41.0 \\
\midrule
\multicolumn{6}{@{}l}{\textit{300 epochs:}} \\
\cdashline{1-7}\\
Seagrass \cite{Ditria2021a}     & 10.8 & 33.6 & 04.0 & 18.8 & 28.0 & 46.4 \\
DeepFish \cite{Saleh2020a}      & 29.8 & 70.0 & 22.4 & 31.8 & 53.0 & 55.6 \\
YouTube-VOS \cite{Xu2018b}      & 15.2 & 33.8 & 14.4 & 23.0 & 32.0 & 40.0 \\
\bottomrule 
\end{tabular}
}
\end{table}

\subsection{Failure Cases} \label{secfailure}

While our model has shown promising results, there are specific scenarios where it fails to perform optimally. 

\begin{itemize}
    \item Occlusion: Our model's performance degrades when several fish are heavily occluded. While it can estimate the fish mask in some parts as long as they are part of the animal body, it struggles when the occlusion is severe, see \cref{fig:11}.

    \item Variability in Fish Size and Shape: The large variability in the size and shape of fish presents a challenge for our model. It can identify a certain shape of fish, but determining the number of fish in an image remains a difficult task.

    \item Segmentation of Foreground Items: Given a set of unlabeled video collections, our model is only capable of segmenting foreground items and cannot distinguish between distinct object instances or semantic classes. Occasionally, the whole object or parts of the object may not be segmented out.

    \item Influence of Training Videos: Our model's performance is highly influenced by the characteristics of training videos, the coverage of object categories, and the motion of both the camera and the objects. This is similar to other data-driven learning techniques.
\end{itemize}

These failure cases provide valuable insights for future improvements to our model.

% %%%%%%%%%%%%%%%%%%%%%%%%%%%%%%%%%%%%%%%%%%%%%%%%%%%%%%%%%%%%%%%%
% \begin{figure*}[h]
% \centering
% \includegraphics[width=0.88\textwidth]{fig_12.png}
% \caption{illustrate a sample of the failure cases of our model. From left, the original image, mask overlay, and the predicted image.
% The images show instances where the model struggles with heavy occlusion, variability in fish size and shape, segmentation of foreground items, and influence of training videos. These scenarios highlight the limitations of our current approach and provide direction for future improvements.}
% \label{fig:12}
% \end{figure*}
% %%%%%%%%%%%%%%%%%%%%%%%%%%%%%%%%%%%%%%%%%%%%%%%%%%%%%%%%%%%%%%%%

%%%%%%%%%%%%%%%%%%%%%%%%%%%%%%%%%%%%%%%%%%%%%%%%%%%%%%%%%%%%%%%%
\section{Discussion} \label{secdisc}

% Fish segmentation and tracking are notoriously difficult tasks, even for human annotators. A robust motion model is necessary to correct the drifting and jittering of the fish. The fish pose may not be consistent among different cameras, causing them to appear upside-down and to move unnaturally. Tracking fish in video data is challenging because their motion is very irregular and small fish may not be visible throughout the entire dataset. The main problem is that segmentation and tracking are time-consuming tasks, especially when dealing with large datasets.

Fish segmentation and tracking are notoriously difficult tasks, especially for small fish in video data where the background, lighting conditions and fish shape can vary significantly. In particular, for real data, the quality of ground truth labels varies from video to video, since it is difficult to annotate the animal's entire path. Therefore, our model aims to generate a pseudo ground truth by leveraging temporal consistency between frames and improving its quality based on self-supervised learning. The key to our proposed model is to leverage the intrinsic temporal consistency between consecutive frames by using the optical flow and background subtraction method to improve the generated labels. This is especially important when the fish is moving quickly and not in the same location in consecutive frames, as is the case in natural data. 
Tracking fish in video data is also challenging because their motion is very irregular and small fish may not be visible throughout the entire dataset. The other problem is that segmentation and tracking are time-consuming tasks, especially when dealing with large datasets.

Our model outperforms the baseline method (the optical flow without background subtraction) with higher $AP$ values in most of the cases. 
% In general, we noticed an improvement of around $0.3$ in the average $AP$ scores. 
Our approach can utilise temporal consistency to produce consistent labels. In the case of the DeepFish dataset \cite{Saleh2020a}, we observed that our proposed unsupervised model results in higher accuracy compared to the Seagrass dataset \cite{Ditria2021a}. This is mainly due to the more challenging videos in the Seagress data set \cite{Ditria2021a} compared to the DeepFish video data \cite{Saleh2020a}.
Furthermore, we show that for different video datasets, our model shows similar accuracy. Therefore, we can expect that the accuracy would be similar when tested under the same conditions but in new underwater video datasets.

In addition, segmentation accuracy does not degrade after training with supervised training, and training converges in only a few epochs, as shown in \cref{tab:abl}.
In our experiments, we found that segmentation quality
has a significant impact on tracking performance. This is because the quality of the produced object bounding box has a high impact on tracking performance. Even in this case, we still achieved decent results.

We also analyzed the robustness of our proposed model with respect to the environmental conditions. We observed degradation of the model's performance when several fish were heavily occluded, like in \cref{fig:11}. However, our proposed model is still able to estimate the fish mask in some parts as long as they are part of the animal body. 
One of the main challenges in this task is the large variability in the size and shape of fish, as well as the variation in the shape of the fish's body. Although it is possible to identify a certain shape of fish, it is not always possible to determine the number of fish in the image.

Given a set of unlabelled video collections, the main limitation of our study is that it is only capable of segmenting foreground items and cannot distinguish between distinct object instances or semantic classes. Occasionally, the entire object or parts of the object may not be segmented out. The performance of our model is highly influenced by the characteristics of training videos, the coverage of object categories, and the motion of both the camera and the objects, similar to other data-driven learning techniques. Our results are based on a few assumptions. One is that a small subset of semantically similar objects (e.g., all fish) exists in the scene, and these objects are likely to share the same motion feature or to be semantically similar. These assumptions are reasonable if the objects are within a certain size range, they all belong to the same class, and most of them share similar colours, shapes, and sizes. Another limitation of our approach is that we used a relatively large number of videos with a relatively small number of object categories (for instance compared to ImageNet). This allows our model to segment objects of all shapes and colours with only a handful of training examples.

One other limitation of our current framework is that in some cases it is unable to detect all the objects that appear in the video. In future work, we intend to study how to develop a detection-based model that is able to detect all objects appearing in a given scene. Therefore, in the next step, we should look for a more robust and generic objectness model that is able to generalise across a variety of object categories and a variety of background types. Further work could be conducted on more fine-grained object segmentation, especially with new video datasets. 

%%%%%%%%%%%%%%%%%%%%%%%%%%%%%%%%%%%%%%%%%%%%%%%%%%%%%%%%%%%%%%%%
\section{Conclusion}  \label{secconc}

In this study, we introduced an innovative unsupervised methodology for the segmentation and tracking of fish in uncontrolled video environments. Our approach leverages a pseudo-label generation method that combines optical flow with background subtraction, followed by an unsupervised refinement network. This method has proven to yield accurate segmentation results when used to train a supervised Deep Neural Network (DNN) for segmentation. Furthermore, our approach has shown its efficacy in tracking applications.

Our methodology was rigorously tested on three challenging datasets, with the results indicating its robustness and adaptability across different scenarios. This suggests that our approach could serve as a valuable tool for researchers and conservationists working with video data in aquatic environments.

Future research directions include extending our methodology to encompass other aquatic species. This extension, however, would necessitate further adaptations to account for the unique movement patterns and physical characteristics of these species. Another promising avenue for future research is the application of our model to autonomous driving systems for tracking-by-detection. Although this application would present additional challenges, such as dealing with faster-moving objects and more complex backgrounds, we believe the core principles of our approach remain applicable.

In conclusion, our study contributes to a significant advancement in the field of video processing for fish behaviour analysis. The proposed methodology not only enhances our ability to study fish behaviour but also has potential implications for conservation efforts by providing more accurate data on fish populations and movements. Despite the limitations and challenges, we believe that our work lays a solid foundation for future research in this area.

\section*{Data availability statement}
Data sets generated and analysed during the current study are publicly available.

\section*{Conflict of interest statement}
The authors declare that they have no known competing financial interests or personal relationships that could have appeared to influence the work reported in this paper.

\section*{Funding}
No funding was received to assist with the preparation of this manuscript.
The authors have no relevant financial or non-financial interests to disclose.

\section*{Ethics approval statement}
This article does not contain any studies with human participants or animals performed by any of the authors. Informed consent was obtained from all individual participants included in the study.

%%%%%%%%%%%%%%%%%%%%%%%%
\clearpage	
% **** bibliography**** 
% \/ \/ \/ \/ 
\ifCLASSOPTIONcaptionsoff
\newpage
\fi
% \IEEEtriggeratref{8}

\bibliographystyle{IEEEtran}
\bibliography{references}

\end{document}